\documentclass[10pt,twocolumn,letterpaper]{article}

\usepackage{iccv}
\usepackage{times}
\usepackage{epsfig}
\usepackage{graphicx}
\usepackage{amsmath}
\usepackage{amssymb}

\usepackage{amsmath,scalerel}
\DeclareMathOperator*{\concat}{\scalerel*{\Vert}{\sum}}
\usepackage{booktabs}
\usepackage{multirow}
\usepackage{pifont}
\usepackage{color}
\usepackage[table]{xcolor}
\usepackage{colortbl}
\usepackage{tabularx}
\usepackage{array} 
\newcolumntype{Y}{>{\centering\arraybackslash}X}

\definecolor{yellow}{rgb}{1,1, 0.6}
\definecolor{orange}{rgb}{1, 0.8, 0.6}

\usepackage{algorithmic}
\usepackage[linesnumbered,ruled,noend]{algorithm2e}
\usepackage{setspace}

\makeatletter
\newcommand{\Statey}{\Statex\hspace*{\ALG@thistlm}}

\usepackage[pagebackref=true,breaklinks=true,letterpaper=true,colorlinks,bookmarks=false]{hyperref}

\usepackage[capitalize]{cleveref}
\crefname{section}{Sec.}{Secs.}
\Crefname{section}{Section}{Sections}
\Crefname{table}{Table}{Tables}
\crefname{table}{Tab.}{Tabs.}
\crefname{algorithm}{Algorithm}{Algorithms}

 \iccvfinalcopy 

\def\iccvPaperID{6102} 

\ificcvfinal\fi

\begin{document}

\title{Hierarchically Decomposed Graph Convolutional Networks for\\Skeleton-Based Action Recognition}

\author{Jungho Lee$^{1}$
,
Minhyeok Lee$^{1}$
,
Dogyoon Lee$^{1}$
,
Sangyoun Lee$^{1}$\\
\vspace{0.01cm}\\
$^{1}$Yonsei University\\
{\tt\small \{2015142131, hydragon516, nemotio, syleee\}@yonsei.ac.kr}
}

\maketitle
\ificcvfinal\thispagestyle{empty}\fi

\begin{abstract}
   Graph convolutional networks (GCNs) are the most commonly used methods for skeleton-based action recognition and have achieved remarkable performance. Generating adjacency matrices with semantically meaningful edges is particularly important for this task, but extracting such edges is challenging problem. To solve this, we propose a hierarchically decomposed graph convolutional network (HD-GCN) architecture with a novel hierarchically decomposed graph (HD-Graph). The proposed HD-GCN effectively decomposes every joint node into several sets to extract major structurally adjacent and distant edges, and uses them to construct an HD-Graph containing those edges in the same semantic spaces of a human skeleton. In addition, we introduce an attention-guided hierarchy aggregation (A-HA) module to highlight the dominant hierarchical edge sets of the HD-Graph. Furthermore, we apply a new six-way ensemble method, which uses only joint and bone stream without any motion stream. The proposed model is evaluated and achieves state-of-the-art performance on four large, popular datasets. Finally, we demonstrate the effectiveness of our model with various comparative experiments. Code is available at \href{https://github.com/Jho-Yonsei/HD-GCN/}{https://github.com/Jho-Yonsei/HD-GCN}.
\end{abstract}

\section{Introduction}
Human action recognition (HAR) is a task that categorizes action classes by receiving video data as input. HAR is used in many applications, such as human–computer interaction and virtual reality. Recently, several RGB-based and skeleton-based HAR methods have been proposed with the development of deep learning technology. However, RGB-based methods~\cite{wang2016temporal,veeriah2015differential} cannot robustly recognize human actions because they are strongly influenced by environmental noises such as background color, brightness of light, and clothing. Therefore, methods using skeleton modality~\cite{yan2018spatial,shi2019two,zhang2020semantics,si2019attention,cheng2020skeleton,cheng2020decoupling,liu2020disentangling,chen2021channel,lee2022leveraging} have attracted attention because they are not affected by these noises. These methods recognize action by receiving 2D or 3D coordinates of major human joints as time-series inputs.

\begin{figure}[t]
	\centering
	\includegraphics[width=\linewidth]{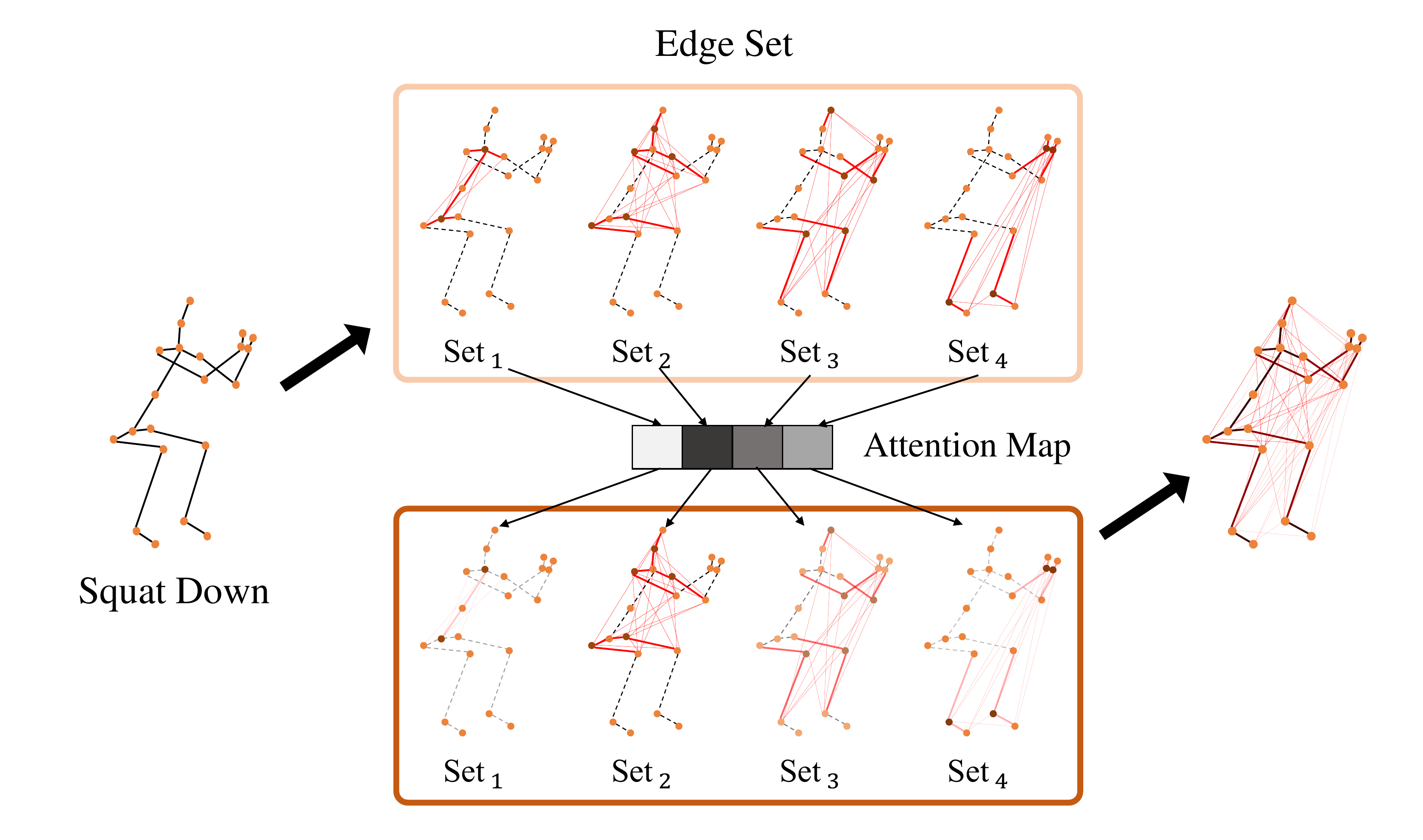}
	\caption{\textbf{The framework of HD-GCN.} The input skeleton is applied with various edge sets through a hierarchically decomposed graph (HD-Graph). The red lines are the edges included in the corresponding hierarchy edge set. The network highlights the major edge sets through the attention map. The darker the color of red line, the more highlighted the edge set, and dotted lines denote unconnected edges.}
	\label{fig:method}
	\vspace{-5mm}
\end{figure}

Recent approaches~\cite{shi2019two,liu2020disentangling,cheng2020decoupling,chen2021channel} have adopted graph convolutional networks (GCNs) to apply human-skeleton graphs to convolutional layers. However, existing GCN-based methods~\cite{yan2018spatial,shi2019two,shi2020skeleton,cheng2020decoupling,chen2021channel} have the following limitations. (1) With the widely used handcrafted graph, the relationships between distant joint nodes are not identified since they use only the relationships of PC edges in the human skeleton. Although the graph with PC edges has a semantic significance, the graph with only PC edges suffers from long-range dependency problem as they are heuristically fixed. However, for humans to recognize actions, relationships between structurally distant joints as well as between adjacent joints are strongly correlated. Although several methods~\cite{shi2019two,chen2021channel} have attempted to solve such limitation by training attention-guided learnable graphs, they still use~\cite{yan2018spatial}'s handcrafted graph with their learnable graphs. Moreover, as the element values of~\cite{yan2018spatial}'s graph are more dominant than those of the learnable graphs, they do not adequately highlight the relationships between distant nodes. (2) Some recent methods~\cite{yan2018spatial,shi2019two,cheng2020decoupling,liu2020disentangling} risk falling into suboptimality by simply aggregating the edge features and ignoring the contribution of each edge, thus incompletely recognizing which edges are significant for each skeleton sample. For example, in the case of a `squat down' action, the interactions between the legs and arms should be highlighted.

Motivated by these limitations, we propose a hierarchically decomposed graph convolutional network (HD-GCN) with a hierarchically decomposed graph (HD-Graph) and attention-guided hierarchy aggregation (A-HA) module. In addition, we present a six-way ensemble method to effectively utilize our HD-Graph. The framework of our proposed methods is shown in~\cref{fig:method} for `squat down' action.

The HD-GCN incorporates GCNs with our HD-Graph, which identifies the relationships between distant joint nodes in the same semantic spaces ($\textit{e}.\textit{g}.$ right and left hands, right and left feet). The same semantic spaces are formed by moving out step by step from the Center of Mass (CoM) node of the graph. For example, if belly is a CoM node, the first semantic space includes the belly node, the next space includes the chest and hip nodes, and the subsequent space includes the left and right shoulder and the left and right hip nodes. The nodes in the same semantic space are defined as hierarchy node set. To detect the relationships between distant joint nodes, network should have large receptive field. The proposed HD-Graph contains both meaningful adjacent and distant joint nodes by connecting all the nodes in neighboring hierarchy node sets and identifies the connectivity between those nodes for large receptive field. We adopt rooted tree-like structure to effectively represent every edges. We apply a spatial edge convolution (S-EdgeConv) layer to consider semantically close edges which cannot be captured by the HD-Graph for each sample. To create the S-EdgeConv layer, we borrow the structure of~\cite{wang2019dynamic}, which is widely used in 3D point clouds. 

To consider the contribution of each edge set, the process of selecting the dominant hierarchical information should depend on the action data sample to give proper attention to the most dominant edge sets. For example, in order to recognize the ``clapping'' action, a hierarchy edge set that includes both hands must be emphasized. To tackle this issue, we propose an attention-guided hierarchy aggregation (A-HA) module, which consists of two submodules: representative spatial average pooling (RSAP) and hierarchical edge convolution (H-EdgeConv). A scaling bias problem occurs if we use the spatial average pooling module without any node extraction process because each node has a different number of adjacent nodes. To prevent this, we apply RSAP, which includes a representative node extraction process that triggers features after the pooling layer to represent each node. To effectively manage hierarchical features obtained by RSAP, we apply a hierarchical edge convolution (H-EdgeConv) layer. The H-EdgeConv treats each hierarchical feature as a graph node and identifies which hierarchical features should be highlighted via the Euclidean distance in feature space. With the RSAP and the H-EdgeConv, our model successfully determines which hierarchy edge sets and joints should be emphasized among the input features.

The existing ensemble method uses four-stream data composed of the joint, bone, joint motion, and bone motion streams, which are the original skeletal coordinates, spatial differential between joint coordinates, and temporal differential of joint, and temporal differential of the bone, respectively. Most existing ensemble methods~\cite{shi2020skeleton,chen2021channel} use additional motion data, but models that solely utilize motion data exhibit relatively inferior performance. To address this problem, we present a new method, a six-way ensemble. We apply this ensemble method by setting three HD-Graphs with joint and bone stream data. Each graph has different CoM nodes to extract features of different semantic spaces (see \textbf{Appendix}).

We conduct extensive experiments on four benchmark action recognition datasets: NTU-RGB+D 60~\cite{shahroudy2016ntu}, NTU-RGB+D 120~\cite{liu2019ntu}, Kinetics-Skeleton~\cite{kay2017kinetics}, and Northwestern-UCLA~\cite{wang2014cross}.

Our main contributions are summarized as follows:
\begin{itemize}
	\item [$\bullet$] We propose a hierarchically decomposed graph (HD-Graph) to thoroughly identify the significant distant edges between the same hierarchy node sets.
	\vspace{-1mm}
	\item [$\bullet$] We propose an attention-guided hierarchy aggregation (A-HA) module to highlight the key edge sets with representative spatial average pooling (RSAP) and hierarchical edge convolution (H-EdgeConv).
	\vspace{-1mm}
	\item [$\bullet$] We use a new six-way ensemble method for skeleton-based action recognition with HD-Graphs that have different center of mass (CoM), which outperforms regular ensemble without any motion data.
	\vspace{-1mm}
	\item [$\bullet$] Our HD-GCN outperforms the state-of-the-arts on four benchmarks for skeleton-based action recognition.
\end{itemize}

\section{Related Work}

\begin{figure*}[t]
	\centering
	\includegraphics[width=\linewidth]{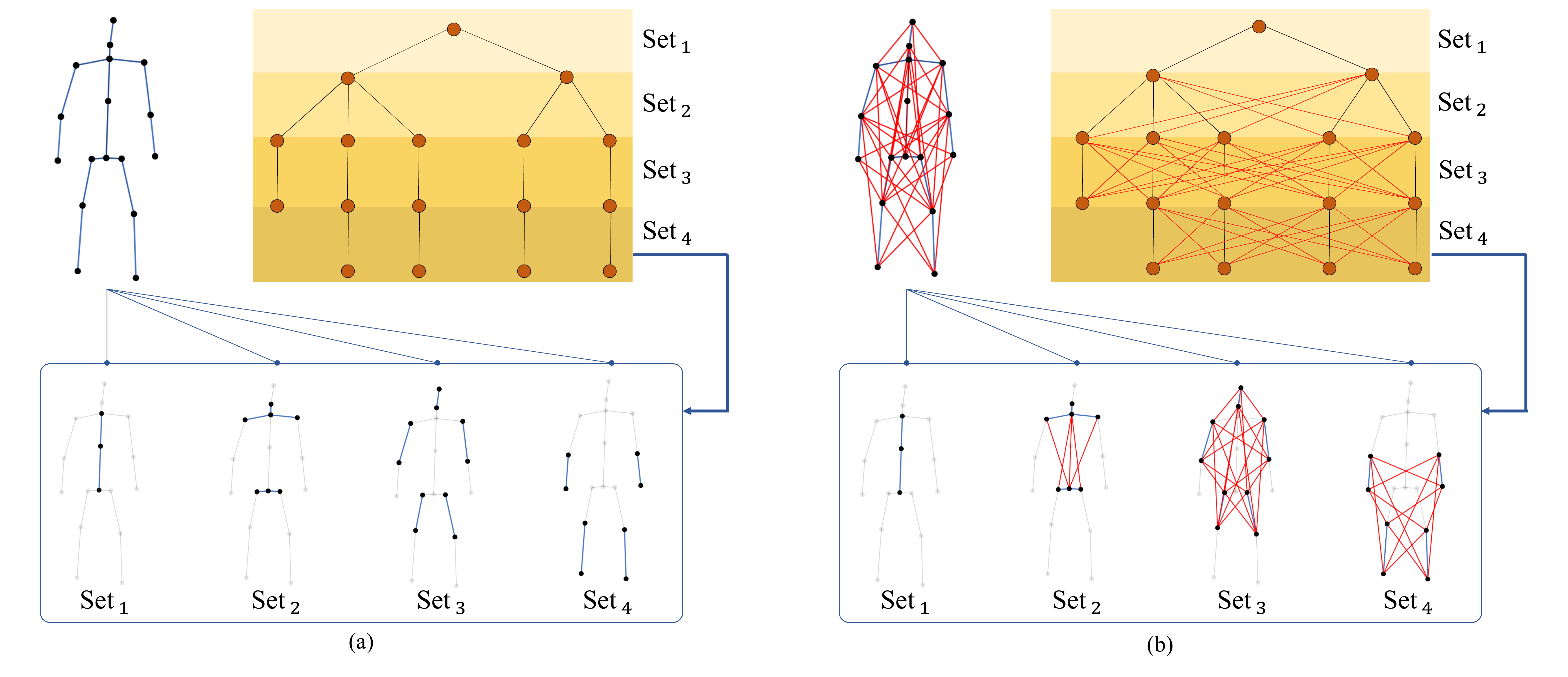}
	\caption{\textbf{(a) Structure of HD-Graph with physically connected (PC) edges.} The human skeleton graph is decomposed into a rooted tree, where PC edges are included in hierarchy sets. \textbf{(b) Structure of HD-Graph with fully connected (FC) edges.} Edges between all nodes in the same semantic space are obtained by connecting all the nodes in adjacent hierarchy edge sets. Blue and red lines stand for PC and FC edges, respectively.}
	\label{fig:hdgraph}
	\vspace{-2mm}
\end{figure*}

\subsection{Action Recognition with GCNs}

In skeleton-based action recognition, human skeletal data are represented by a graph with joint nodes. Most recent approaches use GCN-based methods~\cite{yan2018spatial,shi2019two,cheng2020skeleton,liu2020disentangling,chen2021channel} with ~\cite{yan2018spatial}'s graph structure, which identifies physical connections in the human skeleton. Those GCN-based methods perform remarkably better than methods using handcrafted features~\cite{duvenaud2015convolutional,hamilton2017inductive,kipf2018neural,monti2017geometric,niepert2016learning}. They extract the spatial features representing the relationships between physically connected edges among human skeleton, and they outperform other methods by using them to construct the major relationships between joint nodes in the human skeleton. In particular,~\cite{shi2020skeleton} and~\cite{chen2021channel} propose adaptive attention-based graph structures to learn the sample-wise topological features. However, they might fall into suboptimality because they do not consider the physical prior of the human skeletal structure and allow too much flexibility in network training. To address this issue, we introduce a novel HD-Graph, referencing the known tendencies of human perception

\subsection{Attention Modules for Action Recognition}

The attention mechanism is an essential element for constructing a deep neural network. Using recent attention modules~\cite{hu2018squeeze,woo2018cbam}, networks emphasize important information along a specific dimension. For example, Hu~\etal~\cite{hu2018squeeze} applies channel-wise attention, and Woo~\etal~\cite{woo2018cbam} applies both channel-wise and spatial-wise attentions. These techniques are divided into two categories for GCNs: (1) attention-based graph construction~\cite{shi2019two,chen2021channel} which is a method of forming topologies using a non-local block~\cite{wang2018non} or customized correlation matrices, and (2) spatial-wise, temporal-wise, channel-wise attention, which are commonly used attentions in~\cite{shi2020skeleton,song2022constructing}, and several other networks.

\section{Methodology}

In \cref{sec:hdgraph}, we detail the HD-Graph convolution to solve the problems of the conventional human-skeleton graph~\cite{yan2018spatial}, which includes only PC edges. We also explain the A-HA module in \cref{sec:aha} to highlight dominant hierarchical features. In \cref{sec:ensemble}, we replace the widely used four-stream ensemble method~\cite{shi2020skeleton,chen2021channel,cheng2020decoupling} with a six-way ensemble without motion data streams. Finally, we introduce the HD-GCN, which uses these proposed methods.

\subsection{Preliminaries}\label{sec:preliminary}

\paragraph{\bf{Notations.}} The spatio-temporal graph for human skeleton is represented by $\mathcal{G(V, E)}$, where $\mathcal{V}$ and $\mathcal{E}$ denote the joint and edge groups, respectively. Physically connected edges and fully-connected edges used in~\cref{sec:hdgraph} are denoted as PC-edges and FC-edges, respectively.

\paragraph{\bf{Graph Convolutional Networks.}} 3D time-series skeletal data are represented by $\mathbf{X} \in \mathbb{R}^{3\times T\times V}$, where $V$ and $T$ are the number of joint nodes and the temporal window size, respectively. GCN's operation with input feature map $\mathbf{F}_{in} \in \mathbb{R}^{C\times T\times V}$is as follows:
\begin{equation}
	\label{eq:oldgcn}
	\mathbf{F_{\mathit{out}}}=\sum_{s \in S} \widehat{\mathbf{A}}_{s}\mathbf{F}_{in}\Theta_{s},
\end{equation}
where $\mathit{S}=\{s_{id}, s_{cf}, s_{cp}\}$ denotes graph subsets, and $s_{id}$, $s_{cf}$, and $s_{cp}$ indicate identity, centrifugal, and centripetal joint subsets, respectively. $\Theta_{s}$ denotes the pointwise convolution operation. The normalized adjacency matrix $\widehat{\mathbf{A}}$ is initialized as $\mathbf{\Lambda}^{-\frac{1}{2}} \mathbf{A} \mathbf{\Lambda}^{-\frac{1}{2}} \in \mathbb{R}^{N_{S}\times V \times V}$, where $\mathbf{\Lambda}$ is a diagonal matrix for normalization and $N_{S} = 3$.

\subsection{Hierarchically Decomposed Graph}\label{sec:hdgraph}

Most recent methods have adopted the handcrafted graph proposed by Yan~\etal~\cite{yan2018spatial}, but the HD-Graph is derived through a newly presented method. \cref{fig:hdgraph} shows the framework of the HD-Graph.

\paragraph{\bf{Decomposition into a Rooted Tree.}} The first step is to decompose the graph with PC edges and construct a rooted tree. To decompose a given skeleton into the tree, we need to determine a CoM node, which allows nodes in the same hierarchy edge set to exist in the same semantic space. For example, nodes in the same semantic space, such as elbow and knee joints, or hands and feet, must exist in a hierarchy node set. After choosing the CoM node, the graph is converted into a rooted tree, which includes the hierarchical information of the graph, and defines the directed adjacency matrix $\overrightarrow{\mathbf{A}}_{\textrm{HD}} \in \mathbb{R}^{N_{L} \times V \times V}$ with $N_{L}$ hierarchy layers for $N_{H}$ hierarchy edge sets:
\begin{equation}
	\overrightarrow{\mathbf{A}}_{\textrm{HD}} = \big[\mathcal{E}(H_{1} \to  H_{2}),\  \cdots,\ \mathcal{E}(H_{N_{H}-1} \to H_{N_{H}})\big],
\end{equation}
where $H_{k}$ denotes the $k$-th hierarchy node set and $\mathcal{E}(H_{k} \to H_{k+1})$ denotes a set of edges from $H_{k}$ to $H_{k+1}$. $N_{L}$ and $N_{H}$ are the number of hierarchy layers and hierarchy edge sets, respectively, and $N_{L}=N_{H} - 1$. However, $\overrightarrow{\mathbf{A}}_{\textrm{HD}}$ includes only the directed centrifugal edges. For consistency with existing methods, all the reverse-directed edges from the leaf nodes of the rooted tree in~\cref{fig:hdgraph} to the CoM node must be reflected in the adjacency matrices to cover the centripetal edges. In addition, to get the features of the nodes themselves, the identity edges for each hierarchy node set must be considered. Thus, the adjacency matrices $\overleftrightarrow{\mathbf{A}}_{\textrm{HD}} \in \mathbb{R}^{N_{L} \times N_{S} \times V \times V}$ are defined as follows:
\begin{equation}
	\overleftrightarrow{\mathbf{A}}_{\textrm{HD}} = \big[\mathcal{\ E}_{1},\  \mathcal{E}_{2},\   \cdots ,\   \mathcal{E}_{N_{L}}\ \big],
\end{equation}
\begin{equation}
	\mathcal{E}_{k} = \mathcal{E}(\underbrace{H_{k}\cup H_{k+1}}_{s_{id}},\  \underbrace{H_{k} \to H_{k+1}}_{s_{cp}},\  \underbrace{H_{k+1} \to H_{k}}_{s_{cf}}),
\end{equation}
where $\mathcal{E}_{k}$ denotes the concatenation of the three edge subsets of $\mathit{S}=\{s_{id}, s_{cp}, s_{cf}\}$ and $s_{id}, s_{cp}, s_{cf}$ indicate the identity, centripetal, and centrifugal edge subsets, respectively. Through this construction policy, we create a skeletal graph with bidirectional and identity edges.

\paragraph{\bf{Fully Connected Inter-Hierarchy Edges.}} Decomposed graph $\overleftrightarrow{\mathbf{A}}_{\textrm{HD}}$ has a different number of edge sets from the conventional graph, but the edges are all the same. To identify the relationships between major distant joint nodes, especially those in the same semantic space, we connect all nodes between neighboring hierarchy node sets. In addition, since~\cite{yan2018spatial}'s graph contains the connectivity of only PC edges, not distant relationships, the receptive field is very small with this sparse graph. Applying our fully connected (FC) edges to the rooted tree, the graph becomes denser and makes the receptive field larger than before with more meaningful distant connectivity as shown in \cref{fig:hdgraph} (b). Then, the adjacency matrices are normalized with degree matrices for training stability and we leave all elements of the matrices as learnable parameters for training adaptability.

\paragraph{\bf{HD-Graph Convolution.}} Our HD-Graph convolution includes four parallel branch operations: three graph convolution through HD-Graph and an additional EdgeConv~\cite{wang2019dynamic} operation. To reduce the computational complexity, a linear transformation is applied to all four operations. For three of these operations, our method performs a subset-wise GCN operation in the same way as~\cite{yan2018spatial,shi2019two} for each hierarchy edge set with three edge subsets. However, rather than summing the output values for each subset as in \cref{eq:oldgcn}, we concatenate these output values to the channel dimension:
\begin{equation}
	\mathbf{F}_{\textrm{HD}}^{(k)} = \concat_{s\in S} \left\{\overleftrightarrow{\mathbf{A}}_{\textrm{HD};s}^{(k)} \Phi(\mathbf{F}_{in})\Theta^{(k)}_{s}\right\},
\end{equation}
where $\mathbf{F}_{\textrm{HD}}^{(k)}$ denotes the output feature map of the HD-Graph convolution and function $\Phi$ denotes a linear transformation with parameter $\mathbf{W}\in\mathbb{R}^{C'\times C}$. Note that $\concat$ is a concatenation operation.

\begin{figure}[t]
	\centering
	\includegraphics[width=\linewidth]{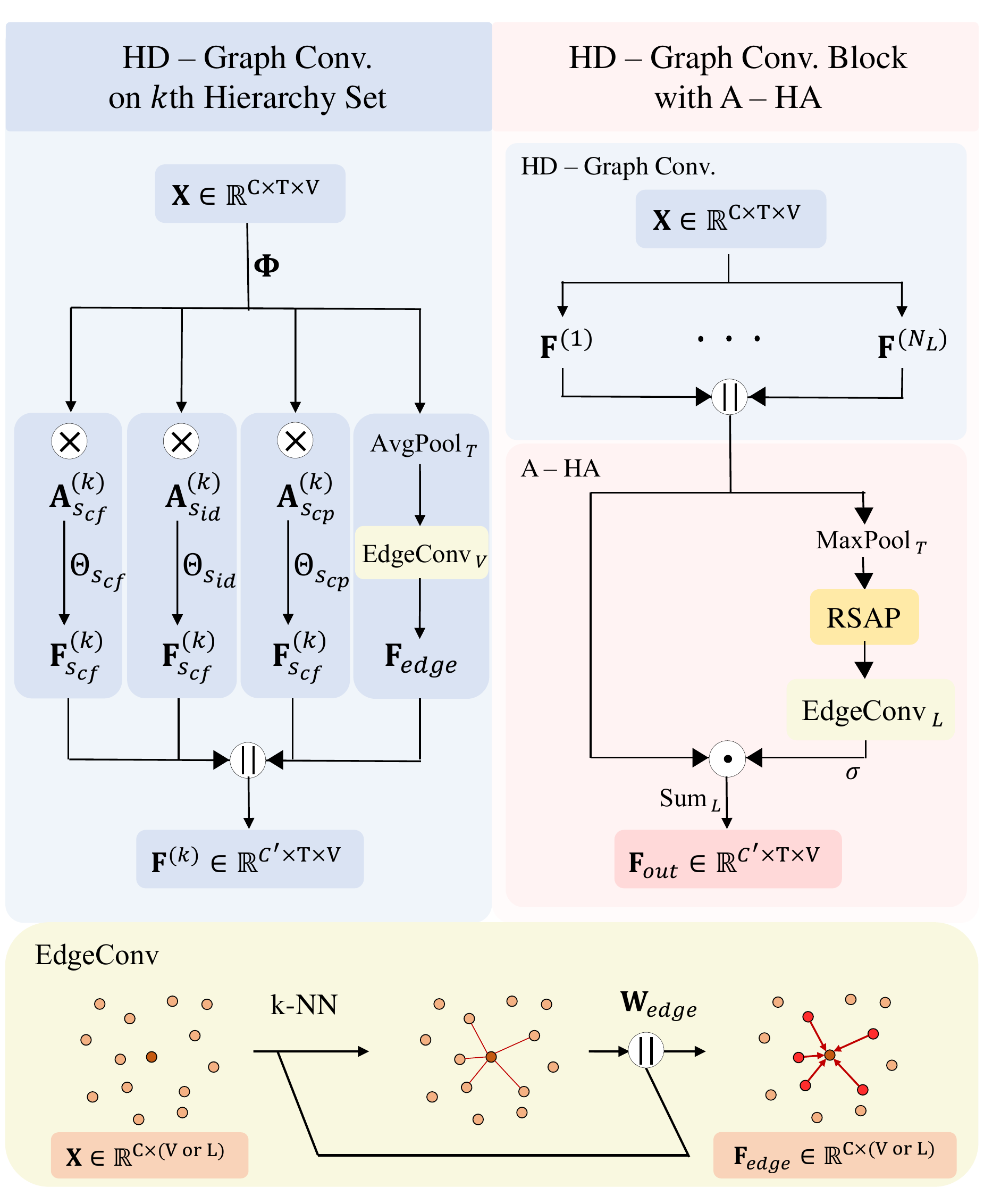}
	\caption{\textbf{HD-Graph convolution operation block with an A-HA module.} The left side shows an operation for one hierarchy edge set, and the right side shows an operation block that concatenates the results for $N_{L}$ edge sets and applies A-HA. The lower part of the figure is EdgeConv, where the EdgeConv subscript indicates the feature space to extract graphical features. The $\times$ and $\cdot$ operations denote matrix and element-wise multiplication.}
	\label{fig:hdgc}
	\vspace{-5mm}
\end{figure}

Although our HD-Graph defines more meaningful node relationships than conventional graph, it may still not be able to extract sample-wise key relationships that reflect the similarities between all nodes in the feature space. To improve this limitation, we adopt EdgeConv~\cite{wang2019dynamic} as the remaining operation, which is used for extracting graphical features through local neighborhood graphs in the feature space. With spatial EdgeConv (S-EdgeConv), our network extracts sample-wise node connectivity, which the HD-Graph does not capture. For our method, S-EdgeConv initially takes the average pooling as the temporal dimension for computational efficiency. Local graphs with local edges are then formed via k-nearest neighbor (k-NN) based on the Euclidean distance, and the local edges as well as identity edges based on the graphs are aggregated via trainable parameters $\mathbf{W}_{edge}\in\mathbb{R}^{C'\times 2C'}$. For the deep neural network, physically close edges are reflected to the initial shallow layers, but as they become deeper, the relationship between semantically similar edges in the feature space are identified and learned. Our whole GCN process is shown in \cref{fig:hdgc} and computed as follows:
\begin{equation}
	\small{
		\mathbf{F}_{\textrm{HD}} \leftarrow \sum_{k=1}^{N_{L}}\left[\mathbf{F}_{\textrm{HD}}^{(k)}\concat z^{(k)}_{\mathcal{V}}\left(\frac{1}{T}\sum_{t=1}^{T}\Phi(\mathbf{F}_{in})\right)\right],
	}
\end{equation}
where $z_{\mathcal{V}}$ denotes the S-EdgeConv operation.

All four branch outputs are concatenated to the channel dimension, with all four computed in the same way for $N_{L}$ hierarchy edge sets. Due to the inherent characteristics of skeletal data, the number of joint nodes included in each dataset is different, and, consequently, the number of hierarchy sets is different. Therefore, we adopt an addition policy for $N_{L}$ hierarchy-wise outputs and a concatenation policy for $N_{S}$ subset-wise outputs. In this way, the dimensionality is maintained, and the common hierarchy-wise aggregation policy is followed for every skeletal dataset by adding all the outputs for different numbers of hierarchical sets.

\subsection{Attention-Guided Hierarchy Aggregation}\label{sec:aha}
The HD-Graph convolution uses an aggregation policy of adding all the hierarchy-wise outputs. However, because each data sample has relationship between specific major edges, we propose an attention-guided hierarchy aggregation (A-HA) module, which applies a weighted-sum policy to the hierarchy dimension with proper attention to the hierarchy-wise outputs. The framework of the HD-Graph convolution with the A-HA module is shown in \cref{fig:hdgc}.

\paragraph{\bf{Representative Spatial Average Pooling.}}
Our A-HA module is applied to feature map $\mathbf{F}_{\textrm{HD}}\in\mathbb{R}^{C\times N_{L}\times T \times V}$ after the HD-Graph convolution. The first step is to extract the temporal frame with the highest score on $\mathbf{F}_{\textrm{HD}}$. RSAP, $\Psi$, is then applied, which is preceded by the extraction of representative nodes in each hierarchy layer. If spatial average pooling is applied without this extraction process, scaling bias occurs because the number of edges connected to each node is different. Therefore, representative node extraction is essential to obtain an appropriate score for attention without any bias. After the extraction, spatial average pooling is applied to hierarchy-wise outputs. Our pooling function $\Psi$ is as follows:
\begin{equation}
	\small{
		\Psi({\mathbf{F}_{\textrm{HD}}^{(k)}}) = \frac{1}{N_{k} + N_{k+1}} \sum_{v \in H_{k}\cup H_{k+1}}\max_{t}\left(\mathbf{F}_{\textrm{HD}}^{(k)}(v)\right),
	}
\end{equation}
where $N_{k}$ denotes the number of vertices in the $H_{k}$ set.

\paragraph{\bf{Hierarchical Edge Convolution.}}
After the RSAP layer, $N_{L}$ hierarchy-wise features in attention feature map $\mathbf{M}$ have not yet shared their information with each other. We treat all $N_{L}$ features as nodes on a graph to learn and reflect similarities in the hierarchical feature space. To apply this process, representative features of these nodes are fed into EdgeConv~\cite{wang2019dynamic}, and the similarities of those nodes are learned based on the Euclidean distance. We also include the self-loop shown in the bottom section of \cref{fig:hdgc} so that the node's own features can be reflected. Our attention map $\mathbf{M}$ operates as follows:
\begin{equation}
	\mathbf{M} = \sigma\left(z_{L}\left(\concat_{k \in L}\left\{\Psi\left(\mathbf{F}_{\textrm{HD}}^{(k)}\right)\right\}\right)\right),
\end{equation}
where $z_{L}$ and $\sigma$ denote H-EdgeConv and the sigmoid function, respectively.

The attention map $\mathbf{M}$ obtained is multiplied by the HD-Graph convolution output feature map $\mathbf{F}_{\textrm{HD}}$, and the output feature map $\mathbf{F}_{out}$ is obtained through a weighted sum to the hierarchy axis as shown in~\cref{fig:hdgc}. Similar to S-EdgeConv in~\cref{sec:hdgraph}, The H-EdgeConv method incorporates the concept of hierarchical edge sets in a physically proximal manner in earlier layers, while in the deeper layers, it emphasizes the presence of semantically similar edge sets. This approach highlights different hierarchical edge sets for each sample and enables the model to learn meaningful representations that capture both physical and semantic properties of the input data.

\subsection{Six-Way Ensemble}\label{sec:ensemble}
Shi~\etal~\cite{shi2019two, shi2020skeleton} have applied a four-stream ensemble method using streams for joints, bones, joint motion, and bone motion. However, as the performances of motion streams are relatively poorer than the performances of joint and bone streams, we adopt an ensemble method with the joint and bone streams without any motion streams. We use three different HD-Graphs, and each graph is used for training with joint and bone streams. The three HD-Graphs have different CoM nodes, which are chest, belly, hip nodes, respectively. In other words, we train joint and bone streams with HD-Graph with the CoM node of chest, and we train the same when the CoM node is belly or hip node. As models with the three different graphs should be trained in different aspects, each of the graphs is composed of different edge sets. For example, if the CoM node is belly, both thigh edges and both upper arm edges are included in the same edge set, whereas when the CoM node is chest, both thigh edges and both forearm edges are included in the same edge set. The details of our six-way ensemble are specified on our \textbf{Appendix}.

\begin{figure}[t]
	\centering
	\includegraphics[width=\linewidth]{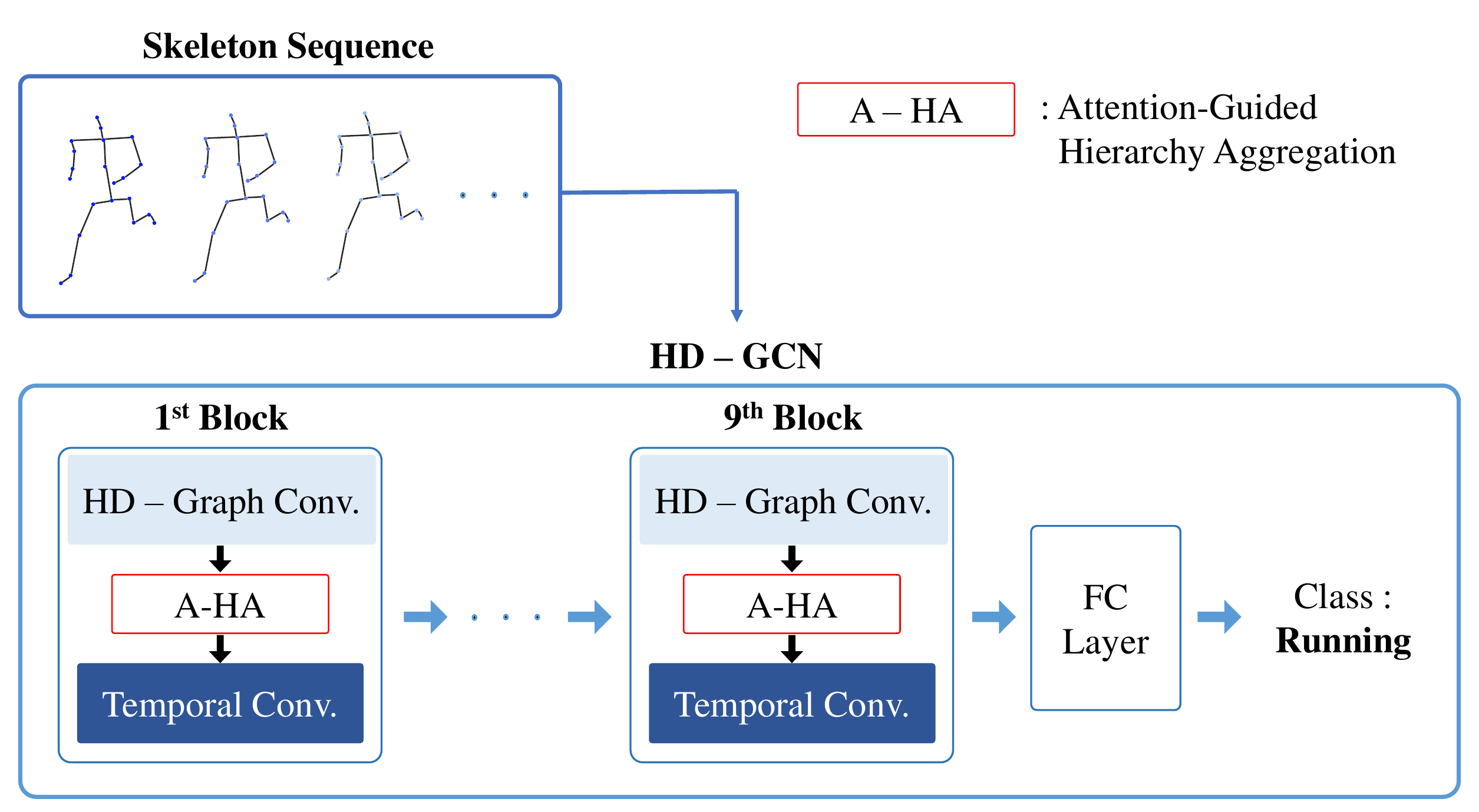}
	\caption{\textbf{The architecture of HD-GCN.} HD-GCN receives the skeleton sequence as input and obtains the class label through nine GCN blocks, and an FC layer, and the softmax function.}
	\label{fig:network}
	\vspace{-5mm}
\end{figure}

\subsection{Network Architecture}\label{sec:architecture}
As shown in \cref{fig:network}, we adopt~\cite{shi2019two} as our baseline network architecture with a total of nine stacked GCN blocks. The numbers of output channels for the blocks are 64, 64, 64, 128, 128, 128, 256, 256, and 256. Each block contains a residual connection~\cite{he2016deep} and is divided into a spatial module, in which the GCN operation proceeds, and a temporal module, which includes the temporal convolutions. Our method use the temporal module of~\cite{chen2021channel}, whose baseline module is~\cite{liu2020disentangling,szegedy2015going}. This module consists of four branch operations. Two are dilated temporal convolutions with kernel size five and dilation one and two, respectively. The remaining branch operations are pointwise convolution and max pooling with kernel size three. Our spatial module consists of an HD-Graph convolution operation and an A-HA module, as introduced in \cref{sec:hdgraph} and \cref{sec:aha}. After passing through all GCN layers with attention to the hierarchy-wise features, the network compresses the feature map through the global average pooling layer and classifies the action sample through the softmax function.

\begin{table*}[]
	{\small	
		\begin{center}
			\resizebox{\textwidth}{!}{
				\begin{tabular}{l|c|c|cc|cc|cc|c}
					\toprule
					\multirow{2}{*}{Methods} & \multirow{2}{*}{Publication} & \multirow{2}{*}{Motion Stream} & \multicolumn{2}{c|}{NTU-RGB+D 60}& \multicolumn{2}{c|}{NTU-RGB+D 120} & \multicolumn{2}{c}{Kinetics-Skeleton} &  \multirow{1}{*}{Northwestern} \\
					& &  &\  X-Sub (\%)    &\  X-View (\%)   &\  X-Sub (\%)   &\  X-Set (\%)  & Top-1 (\%) & Top-5 (\%) & UCLA (\%) \\ \midrule \midrule
					\ ST-GCN~\cite{yan2018spatial}           & AAAI 2018 			 & \ding{55} & 81.5        	 & 88.3       		& 70.7         & 73.2 & 30.7 & 52.8 & -  \\
					\ 2s-AGCN~\cite{shi2019two}          	& CVPR 2019	  		 & \ding{55} & 88.5        	 & 95.1       		& 82.5         & 84.2 & 36.1 & 58.7& -  \\
					\ SGN~\cite{zhang2020semantics}         & CVPR 2020    		 & \ding{51} & 89.0        	 & 94.5       		& 79.2         & 81.5 & - & - & 92.5 \\ 
					\ AGC-LSTM~\cite{si2019attention}		& CVPR 2019			 & \ding{55} & 89.2			& 95.0				& -			& -		& - & - & 93.3  \\
					\ DGNN~\cite{shi2019skeleton}           & CVPR 2019 	    	 & \ding{51} & 89.9         	 & 96.1       		& -            & -    & 36.9 & 59.6 & - \\
					\ Shift-GCN~\cite{cheng2020skeleton}    & CVPR 2020        	 & \ding{51} & 90.7         	 & 96.5       		& 85.9         & 87.6 & - & - & 94.6 \\ 
					\ DC-GCN+ADG~\cite{cheng2020decoupling} & ECCV 2020        	 & \ding{51} & 90.8         	 & 96.6       		& 86.5         & 88.1 & - & - & 95.3 \\ 
					\ DDGCN~\cite{korban2020ddgcn}          & ECCV 2020     		 & \ding{51} & 91.1         	 & \cellcolor{yellow}{97.1} 	  		& -            & -    & 38.1 & 60.8 & - \\         
					\ MS-G3D~\cite{liu2020disentangling}    & CVPR 2020        	 & \ding{55} & 91.5         	 & 96.2       		& 86.9         & 88.4 & 38.0 & 60.9 & - \\    
					\ MST-GCN~\cite{chen2021multi}          & AAAI 2021  			 & \ding{51} & 91.5         	 & 96.6       		& 87.5         & 88.8 & 38.1 & 60.8 & - \\
					\ CTR-GCN~\cite{chen2021channel}        & ICCV 2021    		 & \ding{51} & 92.4         	 & 96.8       		& 88.9         & 90.6 & -  & - & 96.5 \\ 
					\ EfficientGCN-B4~\cite{song2022constructing} & TPAMI 2022  	 & \ding{51} & 91.7         	 & 95.7       		& 88.3         & 89.1 & - & - & - \\ 
					\ STF~\cite{ke2022towards}				& AAAI 2022			 & \ding{55} & 92.5		 	 & 96.9		  		& 88.9	       & 89.9 & 39.9 & - & - \\
					\ InfoGCN (4-ensemble)~\cite{chi2022infogcn}			& CVPR 2022			& \ding{51}	& 92.7			& 96.9				& 89.4			& 90.7		& - & - & 96.6 \\ 
					\ InfoGCN (6-ensemble)~\cite{chi2022infogcn}			& CVPR 2022			& \ding{51}	& \cellcolor{yellow}{93.0}			& \cellcolor{yellow}{97.1}				& \cellcolor{yellow}{89.8}			& \cellcolor{yellow}{91.2}		& -  & - & \cellcolor{yellow}{97.0} \\ 	\midrule
					\ HD-GCN (2-ensemble)     	 &  			 & \ding{55} & 92.4 		 	 & 96.6       		& 89.1   	   & 90.6 & 38.9 & 61.7 & 96.6 \\ 
					\ HD-GCN (4-ensemble)    	& 	 			& \ding{55} & \cellcolor{yellow}{93.0} 		 	 & 97.0       		& \cellcolor{yellow}{89.8}   	   & \cellcolor{yellow}{91.2} & \cellcolor{yellow}{40.3} & \cellcolor{yellow}{63.0}& 96.9 \\ 
					\ \textbf{HD-GCN (6-ensemble)} & & \ding{55} & \cellcolor{orange}{93.4}	 & \cellcolor{orange}{97.2}	& \cellcolor{orange}{90.1} & \cellcolor{orange}{91.6} & \cellcolor{orange}{40.9} & \cellcolor{orange}{63.5}& \cellcolor{orange}{97.2} \\
					\bottomrule
				\end{tabular}
			}
		\end{center}
		\vspace{-2mm}
		\caption{\textbf{Comparisons of the top-1 accuracy (\%) against state-of-the-art methods on the NTU-RGB+D 60, 120, Northwestern-UCLA, and Kinetics-Skeleton datasets.} The \colorbox{orange}{orange} and \colorbox{yellow}{yellow} cells respectively indicate the highest and second-highest value.}
		\label{tab:ntu}
	}
	\vspace{-6mm}
\end{table*}

\section{Experiments}

\subsection{Datasets and Experimental Settings}
\paragraph{\bf{NTU-RGB+D 60.}} NTU-RGB+D 60~\cite{shahroudy2016ntu} is a large dataset used in skeletal action recognition. It contains 56,880 skeleton action samples, performed by 40 different participants and classified into 60 classes. The authors of this dataset recommend two benchmarks. (1) Cross-Subject (X-Sub): 20 of the 40 subjects' actions are used for training, and the remaining 20 are for validation. (2) Cross-View (X-View): Two of the three camera-views are used for training, and the other one is used for validation.

\paragraph{\bf{NTU-RGB+D 120.}} NTU-RGB+D 120~\cite{liu2019ntu} is a dataset in which 57,367 new action samples are added to the NTU-RGB+D 60 dataset. It contains a total of 114,480 skeleton action samples over 120 classes, performed by 106 different subjects. The authors of this dataset recommend two benchmarks: (1) Cross-Subject (X-Sub): 53 of the 106 subjects' actions are used for training, and the remaining 53 are used for validation. (2) Cross-Setup (X-Set): Of the 32 setups, data with even setup IDs are used for training, and the remaining data with odd IDs are used for validation.

\paragraph{\bf{Kinetics-Skeleton.}} The Kinetics-Skeleton dataset is derived from the Kinetics 400 video dataset~\cite{kay2017kinetics}, utilizing the OpenPose pose estimation~\cite{cao2017realtime} to extract 240,436 training and 19,796 testing skeleton sequences across 400 classes. The dataset restricts the number of skeletons per time step to two and eliminates skeletons with lower confidence scores, ensuring high-quality sequences for human action recognition and pose estimation research.
\vspace{-2mm}
\paragraph{\bf{Northwestern-UCLA.}} The Northwestern-UCLA skeleton dataset~\cite{wang2014cross} contains 1494 video clips over 10 classes. Each action is captured through three Kinect cameras with different camera views and is performed by 10 subjects. We adopt the same protocol as NW-UCLA: Two of the three camera-views are used for training, and the other one is used for validation.
\vspace{-2mm}

\paragraph{\bf{Experimental Settings.}} In our experiments, we adopt~\cite{shi2019two} as the backbone. The SGD optimizer is employed with a Nesterov momentum of 0.9 and a weight decay of 0.0004. The number of learning epochs is set to 90, with a warm-up strategy~\cite{he2016deep} applied to the first five epochs for more stable learning. We set the learning rate to decay with cosine annealing~\cite{loshchilov2016sgdr}, with a maximum learning rate of 0.1 and a minimum learning rate of 0.0001. For the NTU-RGB+D datasets, we set the batch size to 64 and use the data preprocessing method from~\cite{zhang2020semantics}. For Kinetics-Skeleton, the batch size is set to 128. In addition, to overcome the absence of belly and hip nodes in the Kinetics-Skeleton, we define the center of both hip joints as CoM hip node, and the center of chest and the hip node as CoM belly node, resulting in a total of 20 nodes. For the Northwestern-UCLA dataset, we set the batch size to 16 and use the data preprocessing method from~\cite{cheng2020skeleton}. All our experiments are conducted on a single RTX 3090 GPU.

\subsection{Comparison with State-of-the-Arts Methods}

Most recent state-of-the-art networks~\cite{shi2020skeleton,cheng2020skeleton,cheng2020decoupling,chen2021channel} adopt a four-way ensemble method, but we adopt the six-way ensemble method described in \cref{sec:ensemble}.

We compare ours with state-of-the-art networks on three datasets: NTU-RGB+D 60~\cite{shahroudy2016ntu}, NTU-RGB+D 120~\cite{liu2019ntu}, Northwestern-UCLA~\cite{wang2014cross}, and Kinetics-Skeleton~\cite{kay2017kinetics}. Comparisons for each dataset are shown in \cref{tab:ntu}. The recognition performance of our HD-GCN has exceeded the state-of-the-arts on every dataset without any motion streams, as shown in \cref{tab:ntu}. With our proposed ensemble method, HD-GCN outperforms the state-of-the-art and shows comparable performance to the 6-way ensemble state-of-the-art using only 4-way ensemble method.

\subsection{Ablation Study}\label{sec:ablation}
In this section, we demonstrate the effectiveness of the proposed HD-GCN. Performance is specified as the cross-subject and cross-setup classification accuracy on the NTU-RGB+D 120~\cite{liu2019ntu} joint stream data.

\begin{table}[h]\scriptsize
	\centering
	\begin{tabularx}{\columnwidth}{X|c|c|Y|Y}
			\toprule
			Graph type & Edges &S-EdgeConv  & X-Sub (\%)   & X-Set (\%)   \\ \midrule \midrule
			Conventional & PC&\ding{55}     		  & 83.5    & 85.4       \\ \midrule
			HD-Graph &           &              & & \\
			\ \ \ A & PC&\ding{55}      & 84.3 ($\uparrow$ 0.8) 	& 86.1 ($\uparrow$ 0.7)        \\
			\ \ \ B & PC&\ding{51}	  & 84.6 ($\uparrow$ 1.1)		& 86.3 ($\uparrow$ 0.9)\\
			\ \ \ C & FC&\ding{55}      & 84.9 ($\uparrow$ 1.4)     & 86.5 ($\uparrow$ 1.1)  \\
			\ \ \ D & FC&\ding{51}      & \bf{85.1} ($\uparrow$ 1.6)   & \bf{86.7} ($\uparrow$ 1.3)       \\ \bottomrule
		\end{tabularx}
	\vspace{-1mm}
	\caption{\textbf{Comparison of the conventional graph and four types of HD-Graph.}}
	\label{tab:hdgraph}
	\vspace{-8mm}
\end{table}

\begin{table}[h]
	\scriptsize
	\centering
	\begin{tabularx}{\columnwidth}{X|c|Y|Y}
		\toprule
		Method &H-EdgeConv& X-Sub (\%)   & X-Set (\%)   \\ \midrule \midrule
		Baseline     		&\ding{55}  & 83.5    & 85.4       \\ \midrule
		HD-GCN		& 	& 		&	\\
		\ \ \ w/o A-HA		 &\ding{55} & 85.1 ($\uparrow$ 1.6)	& 86.7 ($\uparrow$ 1.3) \\
		\ \ \ w/ SAP		&\ding{55} & 85.2 ($\uparrow$ 1.7)	& 86.7 ($\uparrow$ 1.3) \\
		\ \ \ w/ SAP		&\ding{51} & 85.4 ($\uparrow$ 1.9)	& 87.0 ($\uparrow$ 1.6)\\
		\ \ \ w/ RSAP		&\ding{55} & 85.5 ($\uparrow$ 2.0)	& 87.0 ($\uparrow$ 1.6)\\
		\ \ \ w/ RSAP		&\ding{51} & \bf{85.7} ($\uparrow$ 2.2)	& \bf{87.3} ($\uparrow$ 1.9) \\ \bottomrule
	\end{tabularx}
	\vspace{0mm}
	\caption{\textbf{Comparison of various types of attention modules.} SAP denotes the spatial average pooling.}
	\label{tab:aha}
	\vspace{-6mm}
\end{table}

\paragraph{\bf{Hierarchically Decomposed Graph.}}
To proceed with the ablation study for HD-Graph, we set Yan \etal~\cite{yan2018spatial}'s graph as the conventional graph. Here, we use the temporal convolution module of~\cite{liu2020disentangling}, as mentioned in \cref{sec:architecture}, to compare the performance of networks fairly with various graphs. The experimental results are shown in \ref{tab:hdgraph}.

We set the edges of the HD-Graph in different ways to show a gradual performance increase according to the type of graph. There are four main versions of HD-Graph, the first of which is graph A containing only the PC edges. Unlike the conventional graph with one edge set including three fixed subsets, HD-Graph has a flexible number of edge sets, each divided by hierarchy layers with three subsets. Graph B is an extension of A, with the additional operation S-EdgeConv. Graph C contains FC edges for $N_{H}$ hierarchy node sets, and graph D is similar to C but includes S-EdgeConv. The HD-Graph with only PC edges performs better than the conventional graph by a large margin, even though they share the same edges. This proves that it is meaningful to divide the joint nodes by hierarchy edge sets. In addition, the HD-Graph with FC edges and S-EdgeConv performs better for every datasets.

\vspace{-2mm}
\paragraph{\bf{Attention-Guided Hierarchy Aggregation.}}
To prove the effectiveness of the A-HA module, we use a method to change or remove specific parts of our attention module, with the results shown in \cref{tab:aha}. Spatial average pooling (SAP) simply averages along the spatial axis without the representative node extraction process, which performs worse than our RSAP. The poorer performance is due to two factors: (1) scaling bias occurs because the number of nodes in each hierarchy node set is different, and (2) attention through SAP does not represent the corresponding hierarchy node set because it brings the average of the feature vectors of all nodes, not a specific node set. Furthermore, it performs better with H-EdgeConv, which recognizes each hierarchy edge set as a graph node. This proves that because the major edge sets are different for each data sample, it is important to find and highlight edge sets with high similarity based on the Euclidean distance through H-EdgeConv.

\begin{figure}[]
	\centering
	\includegraphics[width=\linewidth]{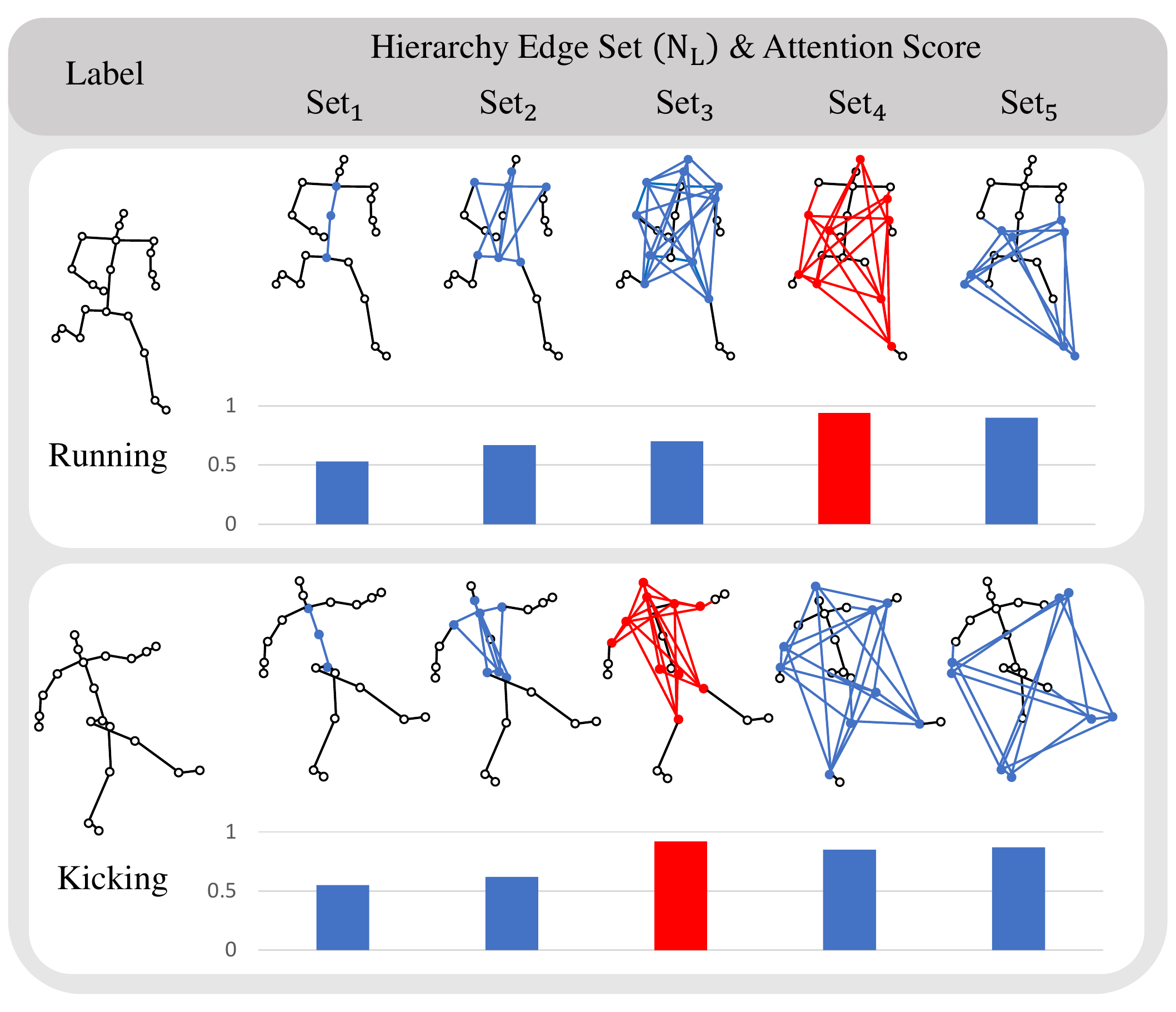}
	\caption{\textbf{Hierarchy-wise attention scores by A-HA for ``Running'' and ``Kicking'' class.} Each score indicates the value of the attention map $\mathbf{M}$ of the A-HA module, and is scaled between 0 and 1 by the sigmoid function.}
	\label{fig:visual}
	\vspace{-5mm}
\end{figure}

The results of the attention score $\mathbf{M}$ of our A-HA module are shown in \cref{fig:visual}. These results show that our module scores edge sets 4 and 5 higher for the ``running'' class, which includes knees and feet, elbows and hands. For the ``Kicking'' class, A-HA gives the highest score to edge set 3, which includes shoulders and hips, followed by edge set 4 and 5. It is reasonable for human visual recognition that the dynamically moving edge set 4, 5 are more important than the stationary and barely moving edge set 3 when running rather than when kicking something.

\paragraph{\bf{Six-Way Ensemble.}}
We use the ensemble method to which three graphs with different CoM nodes are applied, excluding motion streams. \cref{tab:ntu} shows that the HD-GCN with 4-way ensemble outperforms the state-of-the-art~\cite{chi2022infogcn} 4-way methods with motion data and shows comparable performance with the state-of-the-art 6-way method. In addition, when the 6-way ensemble with three different graphs is applied to HD-GCN, it outperforms the state-of-the-art methods. This proves that the features extracted with different CoM nodes are learned in different learning aspects.

\subsection{Comparison of Complexity with Other Models}

Although our model has multiple branch layers for multiple edge sets, it does not cause high complexity because it precedes channel reduction layers. Comparisons of computational complexity with other models are shown in \cref{tab:complexity}. We conduct experiments in the same environment by fixing window size to 64. Our model shows the best performance on NTU-RGB+D 120 joint stream by a large margin even though the computational complexity of our model is the lowest. For multi-stream ensemble, our 4-stream HD-GCN shows almost similar performance to 6-stream InfoGCN~\cite{chi2022infogcn} while having 3.68G fewer FLOPs and 2.70M fewer parameters as shown in~\cref{tab:complexity2}.

\begin{table}[]
	\scriptsize
	\centering
	\begin{tabularx}{\columnwidth}{X|cc|cc}
		\toprule
		& \multicolumn{1}{l}{X-Sub (\%)} & X-Set (\%)    & GFLOPs        & \# Param. (M) \\ \midrule \midrule
		DC-GCN~\cite{cheng2020decoupling} & 84.0$^{\star}$                           &   86.1$^{\star}$            & 2.74          & 3.45          \\
		MS-G3D~\cite{liu2020disentangling}     & 84.9$^{\star}$                               & 86.8$^{\star}$              & 5.22          & 3.22          \\
		CTR-GCN~\cite{chen2021channel}    & 84.9                           & 86.5$^{\star}$              & 1.97          & \textbf{1.46} \\
		InfoGCN~\cite{chi2022infogcn}    & 85.1                           & 86.3          & 1.68          & 1.57          \\ \midrule
		HD-GCN       & \textbf{85.7}                  & \textbf{87.3} & \textbf{1.60} & 1.68        \\ \bottomrule
	\end{tabularx}
	\vspace{0mm}
	\caption{\textbf{Comparison of complexity of the single-stream state-of-the-arts.} ${\star}$: results obtained by the released codes.}
	\label{tab:complexity}
\vspace{-3mm}
\end{table}

\begin{table}[]
	\scriptsize
	\centering
	\begin{tabularx}{\columnwidth}{X|cc|cc}
		\toprule
		& \multicolumn{1}{l}{X-Sub (\%)} & X-Set (\%)    & GFLOPs        & \# Param. (M) \\ \midrule \midrule
		CTR-GCN ${\dagger}$~\cite{chen2021channel}    & 88.9                           & 90.6              & 7.88          & \textbf{5.84} \\
		InfoGCN ${\dagger}$~\cite{chi2022infogcn}    & 89.4                           & 90.7          & 6.72          & 6.28          \\ 
		InfoGCN ${\ddagger}$~\cite{chi2022infogcn}    & 89.8                           & 91.2          & 10.08          & 9.42          \\ \midrule
		HD-GCN ${\dagger}$       & 89.8                 & 91.2 & \textbf{6.40} & 6.72        \\ 
		HD-GCN ${\ddagger}$       & \textbf{90.1}                  & \textbf{91.6} & 9.60 & 10.08        \\ \bottomrule
	\end{tabularx}
	\vspace{0mm}
	\caption{\textbf{Comparison of complexity of the multi-stream state-of-the-arts.} $\dagger$: 4-ensemble, $\ddagger$: 6-ensemble}
	\label{tab:complexity2}
	\vspace{-6mm}
\end{table}

\section{Conclusions}
In this work, we propose a novel hierarchically decomposed graph convolutional network (HD-GCN) for skeleton-based action recognition. We also propose a new framework (HD-Graph) that replaces the existing framework, decomposes all the joint nodes by hierarchy edge sets and considers the connectivity between major distant nodes, which is difficult to identify naturally. We also present an effective attention module (A-HA) composed of representative spatial average pooling (RSAP) layer and hierarchical edge convolution (H-EdgeConv), which applies hierarchy-wise attention for the HD-Graph. In addition, our HD-GCN learns graph-wise features with different patterns through a six-way ensemble method. We derive an effective feature extractor by combining these three methods and empirically verify its effectiveness. Our approach outperforms current state-of-the-art methods on four benchmark datasets.

\clearpage

\appendix
\begin{center}{\bf \Large Appendix}\end{center}
\section{Additional Details of HD-Graph}

\paragraph{\bf{Universality of HD-Graph.}}
It is easier to construct our HD-Graph than existing handcrafted graph~\cite{yan2018spatial} even if ours is composed of more edges than the existing one.~\cite{yan2018spatial}'s graph requires every physically adjacent edges for human joints as shown in~\cref{alg:graph}. On the other hand, our HD-Graph requires only the hierarchy-wise node sets as shown in~\cref{alg:hdgraph}. It verifies that our HD-Graph is more universal than the existing graph in that the requirements of the HD-Graph are fewer than those of the existing one. 

\begin{algorithm} 
	\caption{Physically Adjacent Graph}
	\label{alg:graph}
		\SetAlgoLined
		\KwIn{Physically adjacent inward edge set $\mathcal{E} = \{e_{1}, e_{2}, e_{3}, \cdots, e_{N_{\mathcal{E}}}\}$\\
			NTU-RGB+D : $\mathcal{E}=\{(1, 2), (2, 21), (3, 21), (4, 3), (5, 21), (6, 5), (7, 6),$\\
			$(8, 7), (9, 21), (10, 9), (11, 10), (12, 11), (13, 1),$\\
			$(14, 13), (15, 14), (16, 15), (17, 1), (18, 17), (19, 18),$\\
			$(20, 19), (22, 23), (23, 8), (24, 25), (25, 12)\}$}
		
		Initialize Adjacency matrix $\mathbf{A} \in \mathbb{R}^{3 \times N \times N}$ to $\mathbf{0}$
		
		Assign value of 1 to all diagonal components of $\mathbf{A}^{id}$ to get identity nodes.
		
		\For{$e$ to $\mathcal{E}$}{
			Centripetal edges: $\mathbf{A}^{cp}[e] \gets 1$
			
			Centrifugal edges: $\mathbf{A}^{cf}[reverse(e)] \gets 1$
		}
		
		Initialize degree matrix $\mathbf{\Lambda} \in \mathbb{R}^{3\times N \times N}$ to $\mathbf{0}$
		
		\For{$n=1$ to $N$}{
			$\mathbf{\Lambda}[n, n] \gets$ the number of non-zero elements in column $n$ of $\mathbf{A}$
			
		}
		Normalize adj. matrix with degree matrix:
		$\mathbf{A} \leftarrow \mathbf{\Lambda}^{-\frac{1}{2}} \mathbf{A} \mathbf{\Lambda}^{-\frac{1}{2}}$
		
		\Return {$\mathbf{A}$}
\end{algorithm}

\begin{algorithm}[h]
	\caption{Hierarchically Decomposed Graph}
	\label{alg:hdgraph}
		\SetAlgoLined
		\KwIn{Hierarchy-wise node sets $\mathbf{H} = \{H_{1}, H_{2}, \cdots, H_{L}\}$\\
			NTU-RGB+D : \\
			$H_{1}=\{2\},$\\
			$H_{2}=\{1, 21\},$\\
			$H_{3}=\{13,17,3,5,9\},$\\
			$H_{4}=\{14,18,4,6,10\},$\\
			$H_{5}=\{15,19,7,11\},$\\
			$H_{6}=\{16,20,8,12\},$\\
			$H_{7}=\{22,23,24,25\}$}
		
		Initialize Adjacency matrix $\mathbf{A} \in \mathbb{R}^{(L-1) \times 3 \times N \times N}$ to $\mathbf{0}$
		
		\For{$l = 1$ to $L - 1$}{
			For $H_{l}$ and $H_{l+1}$, include all nodes of those subsets in the diagonal components
			of the adjacency matrix to get identity nodes:
			$\mathbf{A}_{l}^{id}[H_{l}, H_{l}] \gets 1,$
			$\mathbf{A}_{l}^{id}[H_{l+1}, H_{l+1}] \gets 1$
			
			\For{$i=1$ to $length(H_{l})$}{
				\For{$j=1$ to $length(H_{l+1})$}{
					Centripetal edges: $\mathbf{A}_{l}^{cp}[H_{l+1}(j), H_{l}(i)] \gets 1$
					
					Centrifugal edges: $\mathbf{A}_{l}^{cf}[H_{l}(i), H_{l+1}(j)] \gets 1$
				}
			}
			Initialize degree matrix $\mathbf{\Lambda}_{l} \in \mathbb{R}^{3\times N \times N}$ to $\mathbf{0}$
			
			\For{$n=1$ to $N$}{
				$\mathbf{\Lambda}_{l}[n, n] \gets$ the number of non-zero elements in column $n$ of $\mathbf{A}_{l}$
				
			}
			Normalize adj. matrix with degree matrix:
			$\mathbf{A}_{l} \leftarrow \mathbf{\Lambda}_{l}^{-\frac{1}{2}} \mathbf{A}_{l} \mathbf{\Lambda}_{l}^{-\frac{1}{2}}$
		}
		\Return {$\mathbf{A}$}
\end{algorithm}

\begin{figure*}[t]
	\centering
	\includegraphics[width=\linewidth]{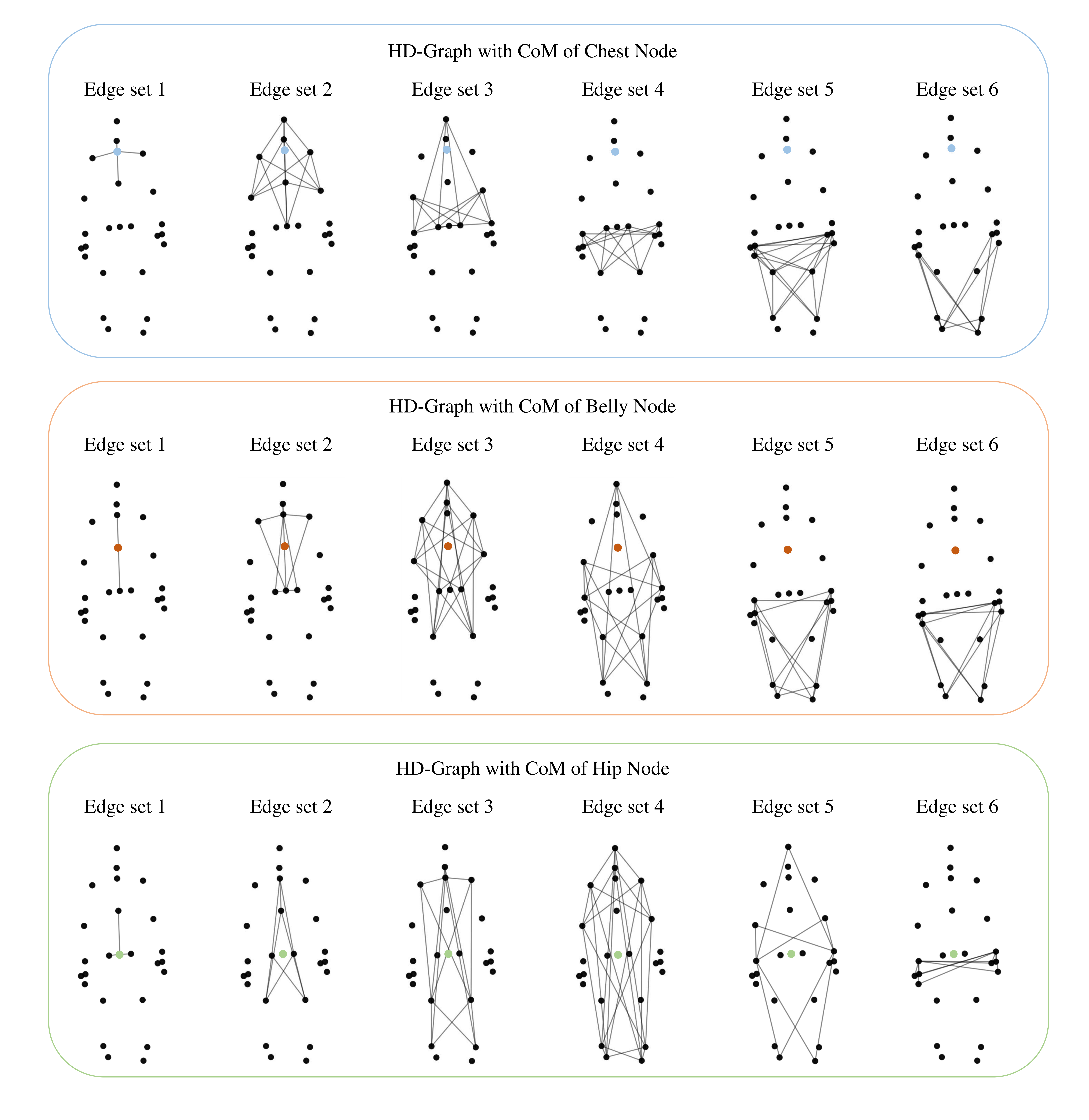}
	\caption{Different HD-Graphs are composed of different edge sets. Colored nodes denote CoM nodes of the graphs.}
	\label{fig:appendix1}
	\vspace{-2mm}
\end{figure*}

\paragraph{\bf{Tree Structures for HD-Graph.}}
Tree structures for skeletal modality has been already proposed in~\cite{liu2016spatio}, which applies depth first search (DFS) algorithm to identify the kinematic dependency relations between the joints. It traverses every joint nodes from the root node to the leaf nodes to model the spatial dependency of the joints. Nevertheless, as~\cite{liu2016spatio}'s tree identifies only the adjacent connections of human joints, it cannot discover direct relationships between structurally distant nodes. Moreover, it is dependent to much on the fixed joint visiting order, which makes the model reflect only the topologically fixed edge features. On the other hand, our HD-Graph is free from those drawbacks. Although we also uses the tree structure to construct the HD-Graph, direct relationships of the structurally distant edges are identified by connecting every nodes for adjacent hierarchy node sets. In addition, because there is no fixed node visiting order in the process of constructing HD-Graph, HD-GCN leverages various edge features via FC-edges and adaptively highlights significant edge sets by A-HA module.

\section{Effectiveness of Six-way Ensemble}
\begin{figure}[t]
	\centering
	\includegraphics[width=\linewidth]{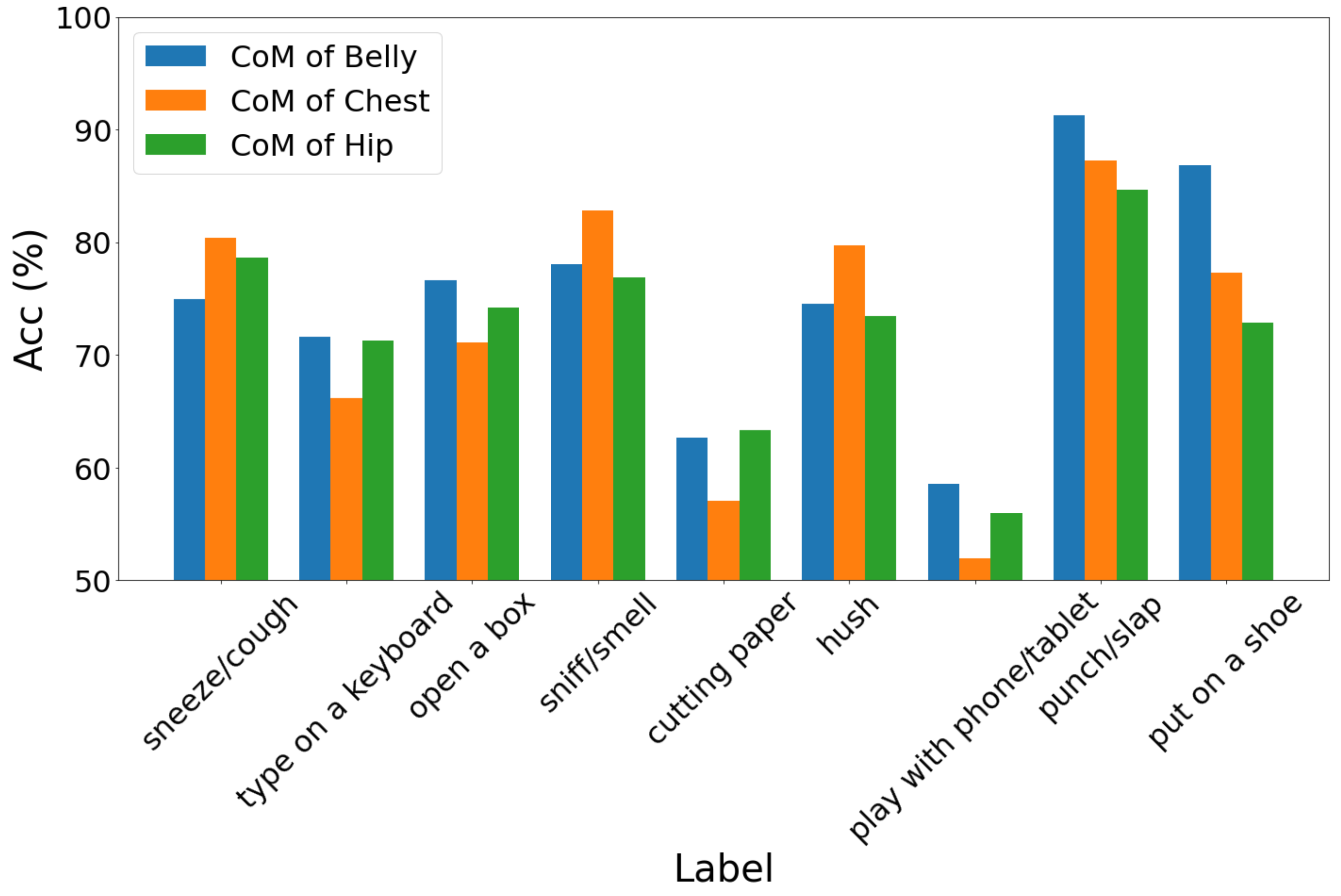}
	\caption{Classification accuracy for several classes with different HD-Graphs.}
	\label{fig:appendix2}
	\vspace{-2mm}
\end{figure}

As we mentioned in our main paper, we propose the ensemble method with joint and bone streams without motion streams. Model with each stream is trained with three different HD-Graphs, which have different CoM nodes; chest, belly, and hip. In other words, training ways for our ensemble methods are as follows: (1) joint stream with CoM of chest node, (2) bone stream with CoM of chest node, (3) joint stream with CoM of belly node, (4) bone stream with CoM of belly node, (5) joint stream with CoM of hip node, (6) bone stream with CoM of hip node. As shown in~\cref{fig:appendix1}, corresponding edge sets for all graphs are different from each others. We compare the performance of those three graphs for several labels on NTU-RGB+D 120 joint dataset as shown in~\cref{fig:appendix2}. The differences between the maximum and minimum accuracy for those three graphs range from 4\% to 13\%. It heuristically proves that the models trained on each of the three graphs have different learning patterns.

\begin{table}[]
	
	\resizebox{\columnwidth}{!}{
		\begin{tabular}{c|c|cc|cc}
			\toprule
			\multirow{2}{*}{CoM}   & \multirow{2}{*}{Stream} & \multicolumn{2}{c|}{NTU-RGB+D 60} & \multicolumn{2}{c}{NTU-RGB+D 120} \\  
			&                         & X-Sub (\%)    & X-View (\%)   & X-Sub (\%)    & X-Set (\%)    \\ \midrule \midrule
			\multirow{2}{*}{Chest} & Joint                   & 90.4              & 95.3          & 85.2             & 87.0           \\
			& Bone                    & 90.7              & 95.2          & 86.7             & 88.1           \\ \midrule
			\multirow{2}{*}{Belly} & Joint                   & 90.5              & 95.6          & 85.7             & 87.3           \\
			& Bone                    & 90.8              & 95.0          & 86.3             & 88.4           \\ \midrule
			\multirow{2}{*}{Hip}   & Joint                   & 90.6              & 95.7          & 85.2             & 87.2           \\
			& Bone                    & 90.9              & 95.1          & 86.4             & 88.4           \\ \midrule
			\multicolumn{2}{c|}{Ensemble}                   & \textbf{93.4}              & \textbf{97.2}          & \textbf{90.1}              & \textbf{91.6}          \\ \bottomrule
		\end{tabular}
	}
\vspace{2mm}
	\caption{Experimental results according to CoMs and data streams.}
	\label{tab:appendix1}
\end{table}

\paragraph{\bf{Ensemble Coefficients.}}
For most recent skeleton-based action recognition models~\cite{chen2021channel,cheng2020decoupling,chi2022infogcn}, optimal coefficients for their ensemble methods should be chosen, which have different values depending on their model. For example,~\cite{chen2021channel} suggests ensemble coefficients of [1.0, 1.0, 0.6, 0.6], which represent joint, bone, joint motion, bone motion streams. Moreover,~\cite{cheng2020decoupling} presents [0.7, 0.7, 0.3, 0.3] and~\cite{cheng2020skeleton} suggests [0.6, 0.6, 0.4, 0.4]. It reduces the universality of the model in that the coefficients should be manually determined. However, there is no need to set those coefficients for six-way ensemble because our method does not require motion streams. Instead of using low-performance motion streams, we use only joint and bone streams and apply the ensemble to all six models with equal contribution. In other words, our ensemble method does not require any ensemble coefficients that determine how much each stream contributes to the model. Applying our ensemble method, our HD-GCN outperforms state-of-the-art methods without the motion streams and manually fixed ensemble coefficients. 

\begin{table}[t]
	{
		\begin{center}
			\resizebox{\columnwidth}{!}{
				\begin{tabular}{l|cc|cc}
					\toprule
					& \multicolumn{1}{l}{X-Sub (\%)} & X-Set (\%)    & GFLOPs        & \# Param. (M) \\ \midrule \midrule
					DC-GCN~\cite{cheng2020decoupling} & 84.0$^{\star}$                           &   86.1$^{\star}$            & 2.74          & 3.45          \\
					MS-G3D~\cite{liu2020disentangling}     & 84.9$^{\star}$                               & 86.8$^{\star}$              & 5.22          & 3.22          \\
					CTR-GCN~\cite{chen2021channel}    & 84.9                           & 86.5$^{\star}$              & 1.97          & \textbf{1.46} \\
					InfoGCN~\cite{chi2022infogcn}    & 85.1                           & 86.3          & 1.68          & 1.57          \\ \midrule
					HD-GCN       & \textbf{85.7}                  & \textbf{87.3} & \textbf{1.60} & 1.68        \\ \bottomrule
				\end{tabular}
			}
		\end{center}
		\caption{\textbf{Comparison of computational and model complexity of the state-of-the-arts.} Each experiment is based on NTU-RGB+D 120 joint stream dataset. ${\star}$ results are implemented based on the released codes.}
		\label{tab:complexityapp}
	}
	\vspace{-4mm}
\end{table}

\paragraph{\bf{Additional Experimental Results.}}
~\cref{tab:appendix1} shows every single experimental result for our six-way ensemble method. Comparing the results of ours in~\cref{tab:appendix1} and other models shown in~\cref{tab:complexityapp}, it shows that our model outperforms the others even on single-stream experiments by a large margin.

\section{Architectures for Kinetics-Skeleton}
\begin{figure}[h]
	\centering
	\includegraphics[width=\linewidth]{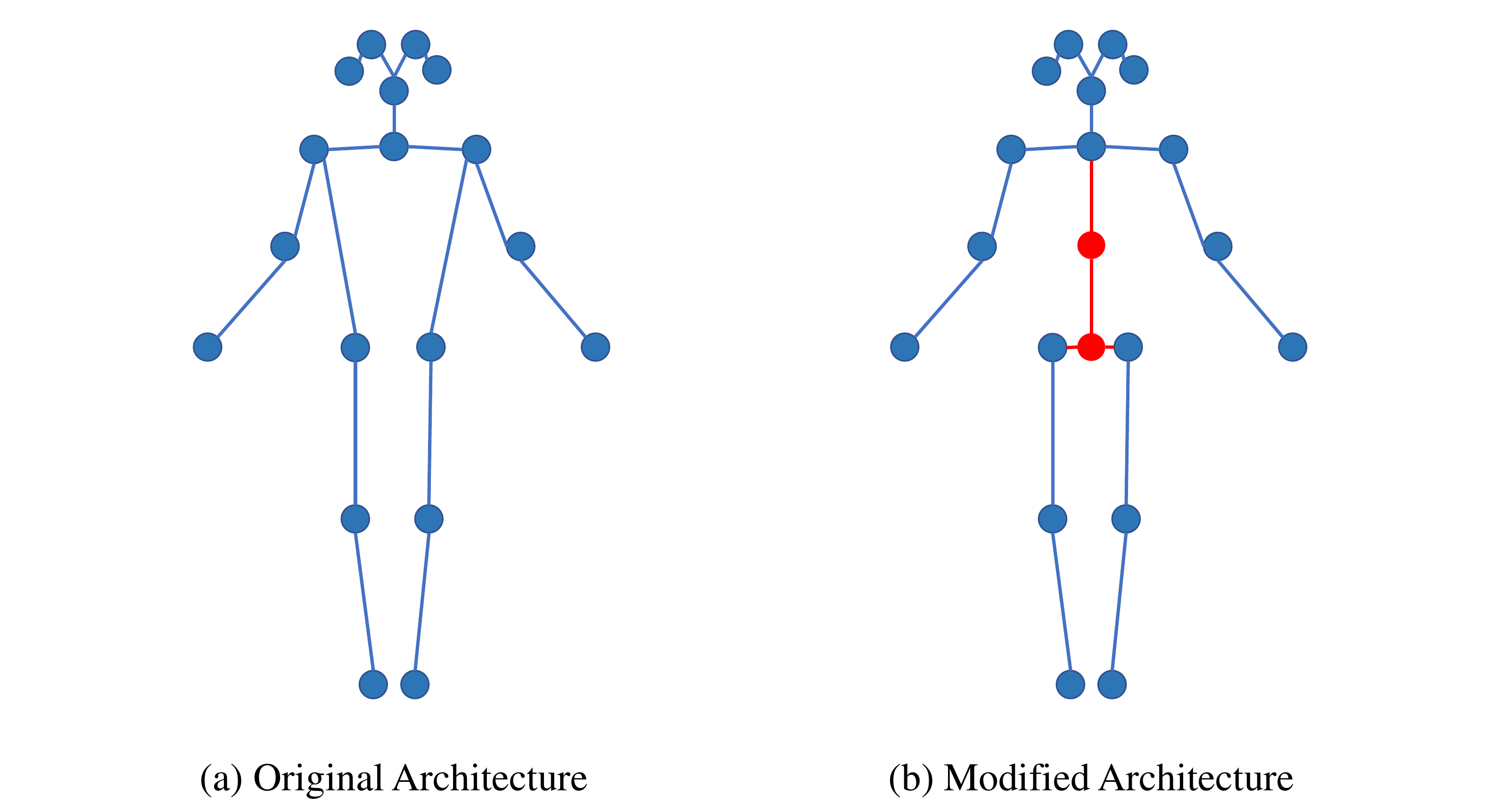}
	\caption{The original and modified version of the Kinetics-Skeleton architecture. The red lines and red circles denote newly generated edges and nodes, respectively.}
	\label{fig:appendix3}
	\vspace{-2mm}
\end{figure}

We modify the original graph of Kinetics-Skeleton~\cite{kay2017kinetics} to apply our HD-Graph. The original architecture of the dataset contains 18 nodes, which does not have hip and belly nodes for CoM, so we manually set those nodes by using existing nodes. Firstly, we set the CoM hip node, which is the middle point of left and right hip nodes. In addition, the belly CoM node is the middle point of chest and hip nodes. The modified skeleton architecture contains 20 nodes due to the generated CoM hip and belly nodes. The original and modified versions of the skeleton are shown in~\cref{fig:appendix3}.

{\small
	\bibliographystyle{ieee_fullname}
	\bibliography{egbib}

\begin{thebibliography}{10}\itemsep=-1pt

\bibitem{cao2017realtime}
Zhe Cao, Tomas Simon, Shih-En Wei, and Yaser Sheikh.
\newblock Realtime multi-person 2d pose estimation using part affinity fields.
\newblock In {\em Proceedings of the IEEE conference on computer vision and
  pattern recognition}, pages 7291--7299, 2017.

\bibitem{chen2021channel}
Yuxin Chen, Ziqi Zhang, Chunfeng Yuan, Bing Li, Ying Deng, and Weiming Hu.
\newblock Channel-wise topology refinement graph convolution for skeleton-based
  action recognition.
\newblock In {\em Proceedings of the IEEE/CVF International Conference on
  Computer Vision}, pages 13359--13368, 2021.

\bibitem{chen2021multi}
Zhan Chen, Sicheng Li, Bing Yang, Qinghan Li, and Hong Liu.
\newblock Multi-scale spatial temporal graph convolutional network for
  skeleton-based action recognition.
\newblock In {\em Proceedings of the AAAI Conference on Artificial
  Intelligence}, volume~35, pages 1113--1122, 2021.

\bibitem{cheng2020decoupling}
Ke Cheng, Yifan Zhang, Congqi Cao, Lei Shi, Jian Cheng, and Hanqing Lu.
\newblock Decoupling gcn with dropgraph module for skeleton-based action
  recognition.
\newblock In {\em Proceedings of the European Conference on Computer Vision
  (ECCV)}, 2020.

\bibitem{cheng2020skeleton}
Ke Cheng, Yifan Zhang, Xiangyu He, Weihan Chen, Jian Cheng, and Hanqing Lu.
\newblock Skeleton-based action recognition with shift graph convolutional
  network.
\newblock In {\em Proceedings of the IEEE/CVF Conference on Computer Vision and
  Pattern Recognition}, pages 183--192, 2020.

\bibitem{chi2022infogcn}
Hyung-gun Chi, Myoung~Hoon Ha, Seunggeun Chi, Sang~Wan Lee, Qixing Huang, and
  Karthik Ramani.
\newblock Infogcn: Representation learning for human skeleton-based action
  recognition.
\newblock In {\em Proceedings of the IEEE/CVF Conference on Computer Vision and
  Pattern Recognition}, pages 20186--20196, 2022.

\bibitem{duvenaud2015convolutional}
David~K Duvenaud, Dougal Maclaurin, Jorge Iparraguirre, Rafael Bombarell,
  Timothy Hirzel, Al{\'a}n Aspuru-Guzik, and Ryan~P Adams.
\newblock Convolutional networks on graphs for learning molecular fingerprints.
\newblock {\em Advances in neural information processing systems}, 28, 2015.

\bibitem{hamilton2017inductive}
Will Hamilton, Zhitao Ying, and Jure Leskovec.
\newblock Inductive representation learning on large graphs.
\newblock {\em Advances in neural information processing systems}, 30, 2017.

\bibitem{he2016deep}
Kaiming He, Xiangyu Zhang, Shaoqing Ren, and Jian Sun.
\newblock Deep residual learning for image recognition.
\newblock In {\em Proceedings of the IEEE Conference on Computer Vision and
  Pattern Recognition}, pages 770--778, 2016.

\bibitem{hu2018squeeze}
Jie Hu, Li Shen, and Gang Sun.
\newblock Squeeze-and-excitation networks.
\newblock In {\em Proceedings of the IEEE conference on computer vision and
  pattern recognition}, pages 7132--7141, 2018.

\bibitem{kay2017kinetics}
Will Kay, Joao Carreira, Karen Simonyan, Brian Zhang, Chloe Hillier, Sudheendra
  Vijayanarasimhan, Fabio Viola, Tim Green, Trevor Back, Paul Natsev, et~al.
\newblock The kinetics human action video dataset.
\newblock {\em arXiv preprint arXiv:1705.06950}, 2017.

\bibitem{ke2022towards}
Lipeng Ke, Kuan-Chuan Peng, and Siwei Lyu.
\newblock Towards to-at spatio-temporal focus for skeleton-based action
  recognition.
\newblock {\em arXiv preprint arXiv:2202.02314}, 2022.

\bibitem{kipf2018neural}
Thomas Kipf, Ethan Fetaya, Kuan-Chieh Wang, Max Welling, and Richard Zemel.
\newblock Neural relational inference for interacting systems.
\newblock In {\em International Conference on Machine Learning}, pages
  2688--2697. PMLR, 2018.

\bibitem{korban2020ddgcn}
Matthew Korban and Xin Li.
\newblock Ddgcn: A dynamic directed graph convolutional network for action
  recognition.
\newblock In {\em European Conference on Computer Vision}, pages 761--776.
  Springer, 2020.

\bibitem{lee2022leveraging}
Jungho Lee, Minhyeok Lee, Suhwan Cho, Sungmin Woo, and Sangyoun Lee.
\newblock Leveraging spatio-temporal dependency for skeleton-based action
  recognition.
\newblock {\em arXiv preprint arXiv:2212.04761}, 2022.

\bibitem{liu2019ntu}
Jun Liu, Amir Shahroudy, Mauricio Perez, Gang Wang, Ling-Yu Duan, and Alex~C
  Kot.
\newblock Ntu rgb+ d 120: A large-scale benchmark for 3d human activity
  understanding.
\newblock {\em IEEE Transactions on Pattern Analysis and Machine Intelligence},
  42(10):2684--2701, 2019.

\bibitem{liu2016spatio}
Jun Liu, Amir Shahroudy, Dong Xu, and Gang Wang.
\newblock Spatio-temporal lstm with trust gates for 3d human action
  recognition.
\newblock In {\em European conference on computer vision}, pages 816--833.
  Springer, 2016.

\bibitem{liu2020disentangling}
Ziyu Liu, Hongwen Zhang, Zhenghao Chen, Zhiyong Wang, and Wanli Ouyang.
\newblock Disentangling and unifying graph convolutions for skeleton-based
  action recognition.
\newblock In {\em Proceedings of the IEEE/CVF Conference on Computer Vision and
  Pattern Recognition}, pages 143--152, 2020.

\bibitem{loshchilov2016sgdr}
Ilya Loshchilov and Frank Hutter.
\newblock Sgdr: Stochastic gradient descent with warm restarts.
\newblock {\em arXiv preprint arXiv:1608.03983}, 2016.

\bibitem{monti2017geometric}
Federico Monti, Davide Boscaini, Jonathan Masci, Emanuele Rodola, Jan Svoboda,
  and Michael~M Bronstein.
\newblock Geometric deep learning on graphs and manifolds using mixture model
  cnns.
\newblock In {\em Proceedings of the IEEE conference on computer vision and
  pattern recognition}, pages 5115--5124, 2017.

\bibitem{niepert2016learning}
Mathias Niepert, Mohamed Ahmed, and Konstantin Kutzkov.
\newblock Learning convolutional neural networks for graphs.
\newblock In {\em International conference on machine learning}, pages
  2014--2023. PMLR, 2016.

\bibitem{shahroudy2016ntu}
Amir Shahroudy, Jun Liu, Tian-Tsong Ng, and Gang Wang.
\newblock Ntu rgb+ d: A large scale dataset for 3d human activity analysis.
\newblock In {\em Proceedings of the IEEE Conference on Computer Vision and
  Pattern Recognition}, pages 1010--1019, 2016.

\bibitem{shi2019skeleton}
Lei Shi, Yifan Zhang, Jian Cheng, and Hanqing Lu.
\newblock Skeleton-based action recognition with directed graph neural
  networks.
\newblock In {\em Proceedings of the IEEE/CVF Conference on Computer Vision and
  Pattern Recognition}, pages 7912--7921, 2019.

\bibitem{shi2019two}
Lei Shi, Yifan Zhang, Jian Cheng, and Hanqing Lu.
\newblock Two-stream adaptive graph convolutional networks for skeleton-based
  action recognition.
\newblock In {\em Proceedings of the IEEE/CVF Conference on Computer Vision and
  Pattern Recognition}, pages 12026--12035, 2019.

\bibitem{shi2020skeleton}
Lei Shi, Yifan Zhang, Jian Cheng, and Hanqing Lu.
\newblock Skeleton-based action recognition with multi-stream adaptive graph
  convolutional networks.
\newblock {\em IEEE Transactions on Image Processing}, 29:9532--9545, 2020.

\bibitem{si2019attention}
Chenyang Si, Wentao Chen, Wei Wang, Liang Wang, and Tieniu Tan.
\newblock An attention enhanced graph convolutional lstm network for
  skeleton-based action recognition.
\newblock In {\em Proceedings of the IEEE/CVF Conference on Computer Vision and
  Pattern Recognition}, pages 1227--1236, 2019.

\bibitem{song2022constructing}
Yi-Fan Song, Zhang Zhang, Caifeng Shan, and Liang Wang.
\newblock Constructing stronger and faster baselines for skeleton-based action
  recognition.
\newblock {\em IEEE Transactions on Pattern Analysis and Machine Intelligence},
  2022.

\bibitem{szegedy2015going}
Christian Szegedy, Wei Liu, Yangqing Jia, Pierre Sermanet, Scott Reed, Dragomir
  Anguelov, Dumitru Erhan, Vincent Vanhoucke, and Andrew Rabinovich.
\newblock Going deeper with convolutions.
\newblock In {\em Proceedings of the IEEE conference on computer vision and
  pattern recognition}, pages 1--9, 2015.

\bibitem{veeriah2015differential}
Vivek Veeriah, Naifan Zhuang, and Guo-Jun Qi.
\newblock Differential recurrent neural networks for action recognition.
\newblock In {\em Proceedings of the IEEE international conference on computer
  vision}, pages 4041--4049, 2015.

\bibitem{wang2014cross}
Jiang Wang, Xiaohan Nie, Yin Xia, Ying Wu, and Song-Chun Zhu.
\newblock Cross-view action modeling, learning and recognition.
\newblock In {\em Proceedings of the IEEE Conference on Computer Vision and
  Pattern Recognition}, pages 2649--2656, 2014.

\bibitem{wang2016temporal}
Limin Wang, Yuanjun Xiong, Zhe Wang, Yu Qiao, Dahua Lin, Xiaoou Tang, and
  Luc~Van Gool.
\newblock Temporal segment networks: Towards good practices for deep action
  recognition.
\newblock In {\em European conference on computer vision}, pages 20--36.
  Springer, 2016.

\bibitem{wang2018non}
Xiaolong Wang, Ross Girshick, Abhinav Gupta, and Kaiming He.
\newblock Non-local neural networks.
\newblock In {\em Proceedings of the IEEE conference on computer vision and
  pattern recognition}, pages 7794--7803, 2018.

\bibitem{wang2019dynamic}
Yue Wang, Yongbin Sun, Ziwei Liu, Sanjay~E Sarma, Michael~M Bronstein, and
  Justin~M Solomon.
\newblock Dynamic graph cnn for learning on point clouds.
\newblock {\em Acm Transactions On Graphics (tog)}, 38(5):1--12, 2019.

\bibitem{woo2018cbam}
Sanghyun Woo, Jongchan Park, Joon-Young Lee, and In~So Kweon.
\newblock Cbam: Convolutional block attention module.
\newblock In {\em Proceedings of the European conference on computer vision
  (ECCV)}, pages 3--19, 2018.

\bibitem{yan2018spatial}
Sijie Yan, Yuanjun Xiong, and Dahua Lin.
\newblock Spatial temporal graph convolutional networks for skeleton-based
  action recognition.
\newblock In {\em Thirty-second AAAI Conference on Artificial Intelligence},
  2018.

\bibitem{zhang2020semantics}
Pengfei Zhang, Cuiling Lan, Wenjun Zeng, Junliang Xing, Jianru Xue, and Nanning
  Zheng.
\newblock Semantics-guided neural networks for efficient skeleton-based human
  action recognition.
\newblock In {\em Proceedings of the IEEE/CVF Conference on Computer Vision and
  Pattern Recognition}, pages 1112--1121, 2020.

\end{thebibliography}
}

\end{document}